\documentclass[letterpaper, 10 pt, conference]{ieeeconf}  

\IEEEoverridecommandlockouts
\pdfoutput=1

\usepackage{xparse} 
\usepackage{amsmath} 

\providecommand{\norm}[1]{\lVert#1\rVert}
\usepackage{amssymb}
\usepackage{amsfonts} 
\usepackage{pgfplots}
\usepackage{bbm}
\usepackage{color}
\usepackage{tikz}
\usetikzlibrary{plotmarks}
\usetikzlibrary{bayesnet}

\usepackage{graphicx}
\usepackage{cleveref}
\usepackage{subcaption}
\usepackage[noadjust]{cite}

\usepackage{url}
\usepackage{framed}

\usepackage{algpseudocode}
\usepackage{algorithm}

\usepackage{threeparttable}
\usepackage{booktabs}
\usepackage{arydshln}

\usepackage{balance}
\usepackage{enumerate}   

\usepackage{xparse}

\NewDocumentCommand\bbm{}{ \begin{bmatrix} }
\NewDocumentCommand\ebm{}{ \end{bmatrix} }

\definecolor{pyplotred}{RGB}{255, 127, 14}
\definecolor{pyplotblue}{RGB}{31, 119, 180	}
\definecolor{pyplotgreen}{RGB}{44, 160, 44	}
\definecolor{mygray}{rgb}{0.4,0.4,0.4}

\newcommand\oc[1]{}

\title{\LARGE \bf
Manipulability Maximization Using\\ Continuous-Time Gaussian Processes}
\author{Filip Mari\'c$^{* \dagger}$, Oliver Limoyo$^*$, Luka Petrovi\'c$^{\dagger}$, Ivan Petrovi\'c$^{\dagger}$, and Jonathan Kelly$^*$
\thanks{This research was supported in part by a Dean's Catalyst Professorship from the University of Toronto and the European Regional Development Fund under the grant KK.01.1.1.01.0009 (DATACROSS).}
\thanks{ $^*$ Filip Mari\'c, Oliver Limoyo, and Jonathan Kelly  are with the University of Toronto, Institute for Aerospace Studies, Space and Terrestrial Autonomous Robotic Systems laboratory, Canada. \{\texttt{<first name>.<last name>@robotics.utias.utoronto.ca}\}}
\thanks{ $^\dagger$ Filip Mari\'c, Luka Petrovi\'c, and Ivan Petrovi\'c are with the University of Zagreb, Faculty of Electrical Engineering and Computing, Laboratory for Autonomous Systems and Mobile Robotics, Croatia. \{\texttt{<first name>.<last name>@fer.hr}\}}
}
\begin{document}

\maketitle
\thispagestyle{empty}
\pagestyle{empty}

\begin{abstract}
A significant challenge in motion planning is to avoid being in or near \emph{singular configurations} (\textit{singularities}), that is, joint configurations that result in the loss of the ability to move in certain directions in task space. A robotic system's capacity for motion is reduced even in regions that are in close proximity to (i.e., neighbouring) a singularity.
In this work we examine singularity avoidance in a motion planning context, finding trajectories which minimize proximity to singular regions, subject to constraints.
We define a manipulability-based likelihood associated with singularity avoidance over a continuous trajectory representation, which we then maximize using a \textit{maximum a posteriori} (MAP) estimator.
Viewing the MAP problem as inference on a factor graph, we use gradient information from interpolated states to maximize the trajectory's overall manipulability.
Both qualitative and quantitative analyses of experimental data show increases in manipulability that result in smooth trajectories with visibly more dexterous arm configurations.
\end{abstract}

\section{Introduction}
\label{section:introduction}
\begin{figure}
	\vspace{0.7\baselineskip}
	\centering
	\begin{subfigure}[]{0.49\columnwidth}
		\includegraphics[width=\textwidth]{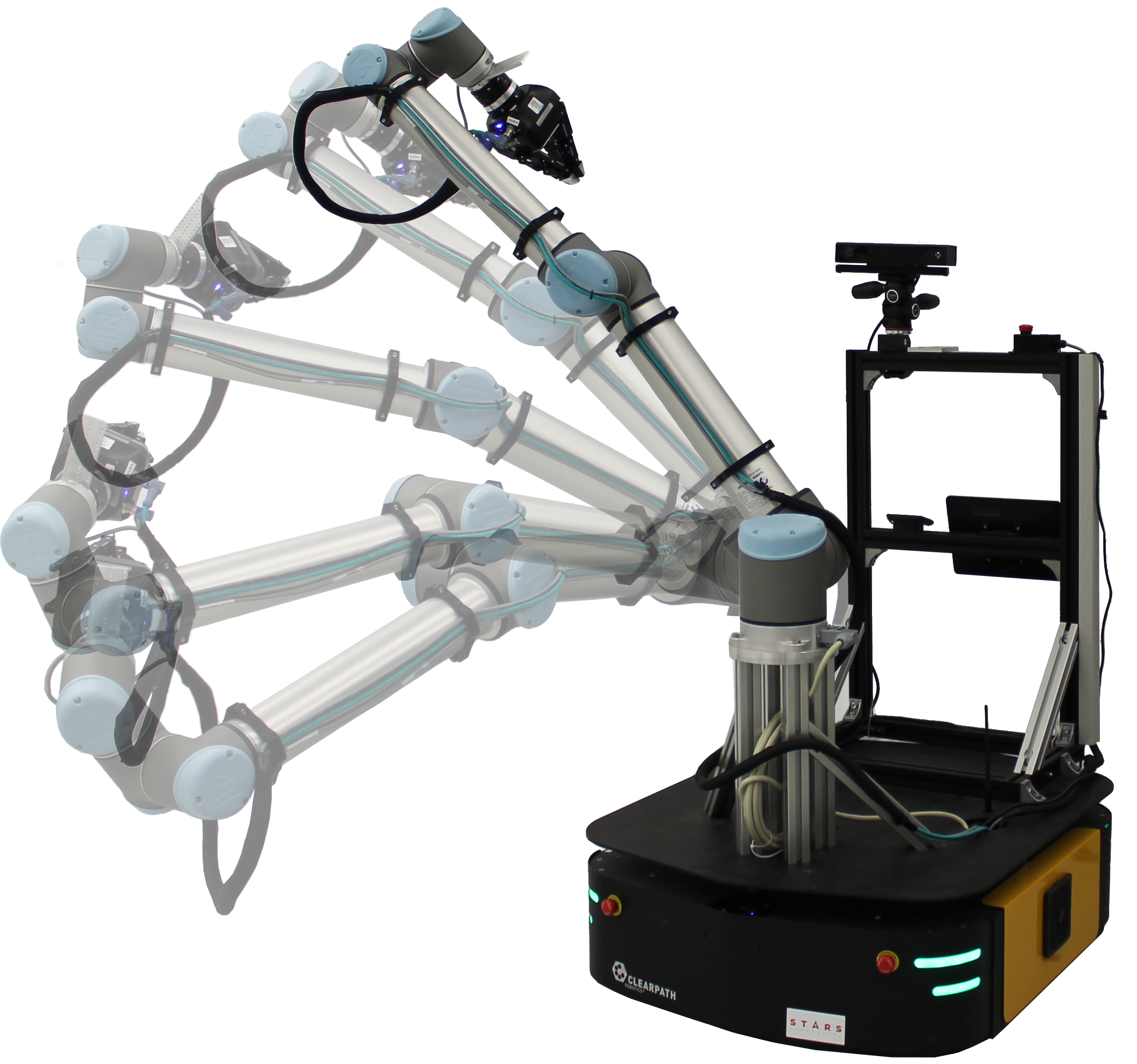}
	\end{subfigure}
	\begin{subfigure}[]{0.49\columnwidth}
		\includegraphics[width=\textwidth]{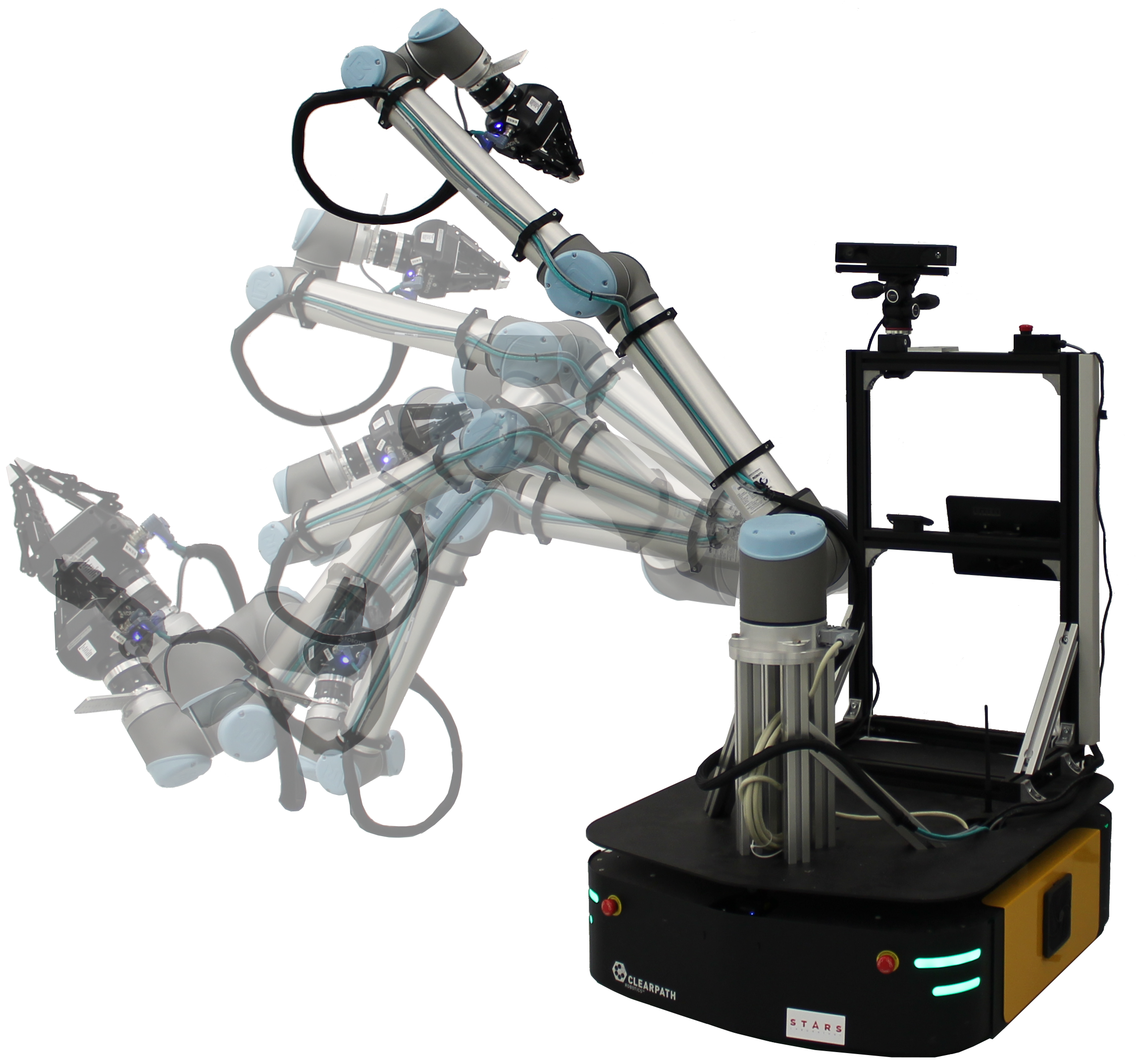}
	\end{subfigure}
	\vspace{2mm}
	\caption{Comparison of two solutions for reaching a 3D point (position) goal from a given near-singular starting configuration (caused by fully extending the arm).
	The left image shows a solution based purely on inverse kinematics, which maintains low manipulability throughout.
	The image on the right shows a trajectory generated by our method, which attempts to avoid excessive arm extension.}
	\label{fig:IRL}
\end{figure}

Motion planning is a fundamental challenge for robotic systems that must execute complex tasks. 
It is possible for motion planning methods to produce trajectories requiring large joint velocities in response to small changes in task space constraints, particularly when na\"{i}ve sampling-based initialization is used.
The goal of singularity avoidance (see below) in motion planning is to generate trajectories avoiding such configurations, known as \textit{singularities}. The arm trajectory shown on the left side of Fig.\ \ref{fig:IRL} is an example in which the configuration is initially (and throughout the motion) nearly singular, with the arm fully extended. On the right side of Fig.\ \ref{fig:IRL} is a trajectory that results in the same final 3D position of the end effector in task space, but that avoids these near-singular configurations throughout.

A configuration's proximity to a singularity can be inferred using the manipulability ellipsoid \cite{sciavicco2012modelling}, which gives a measure of the robot's capacity to perform task space motions.
The manipulability measure introduced by Yoshikawa in \cite{yoshikawa1985manipulability} is proportional to this ellipsoid's volume and has previously been used for singularity avoidance in motion planning.
Velocity-level redundancy resolution \cite{chiaverini1997singularity}, search \cite{guilamo2006manipulability}, and optimization \cite{dufour2017integrating} methods have all been used to maximize the manipulability measure of a single configuration.
In this paper we propose a novel method that considers singularity avoidance as an optimization problem over a continuous trajectory representation.

By choosing to represent the robot's trajectory as a sample from a continuous-time Gaussian process \cite{barfoot2014batch}, the above singularity avoidance problem can be formulated as probabilistic inference; a maximum a posteriori (MAP) estimator can be used to find a solution that is, locally, relatively far from singular regions \cite{gpmp-ijrr}.
By formulating this problem as inference on a factor graph \cite{dong2016motion}, we can interpolate over the trajectory to provide additional gradient information.
Further, replanning on a factor graph \cite{dong2016motion} can be performed in a very efficient manner using the Smoothing and Mapping (SAM) family of algorithms \cite{dellaert2006square}.
This results in a smooth trajectory which maintains high manipulability.
Moreover, the resulting trajectory can be queried at any point, allowing us to monitor the robot's manipulability throughout.
We make the following contributions herein:
\begin{enumerate}[(i)]
\itemsep4pt
	\item we formulate the problem of singularity avoidance using a continuous-time Gaussian process trajectory representation,
	\item we define the likelihood of a given configuration \emph{not} being singular using a known manipulability measure,
	maximizing this measure over the entire trajectory using a MAP estimator, and
	\item we demonstrate that the likelihood gradient information from interpolated states can be used to further improve the resulting trajectories.
\end{enumerate}

\section{Singularity avoidance}
\label{section:singularities}
Consider a joint configuration $\boldsymbol{\theta}_i$ as the state of a trajectory $\boldsymbol{\theta}$ at time $\tau_i$.
The kinematic relationship between configuration and task space velocities at $\boldsymbol{\theta}_i$ for an n-DOF robot is defined as
\vspace{-3mm}
\begin{align}
\label{eq:jacobian}
\dot{\mathbf{x}} = \mathbf{J}\left(\boldsymbol{\theta}_i\right)\boldsymbol{\omega} ,
\end{align}
where $\mathbf{J}\left( \boldsymbol{\theta}_i\right) \in \mathbb{R}^{m\times n}$ is the robot Jacobian matrix at $\boldsymbol{\theta}_i$, while $\boldsymbol{\omega} \in \mathbb{R}^{n}$ and $\dot{\textbf{x}} \in \mathbb{R}^{m}$ are the configuration and task space velocities at $\tau_i$, respectively.
Now, consider an $n$-dimensional ellipsoid in the space of unit joint velocities $\|\boldsymbol{\omega}\|^2 = 1$; we can define the mapping to the Cartesian (task) velocity space as
\begin{align}\label{eq:ellipsoid}
\|\boldsymbol{\omega}\|^2 = \dot{\mathbf{x}}^T\left(\mathbf{J}\mathbf{J}^T\right)^{-1}\dot{\mathbf{x}}.
\end{align}
From Eq.\ \eqref{eq:ellipsoid}, we see that the scaling of joint velocities to the task space depends on the conditioning of the positive semi-definite matrix $\mathbf{J}\mathbf{J}^T$.
Configurations that result in the matrix $\mathbf{J}\mathbf{J}^T$ being non-invertible are termed \textit{singularities}.

\subsection{Manipulability}
\begin{figure}
\centering
    \def\svgwidth{0.6\columnwidth}
    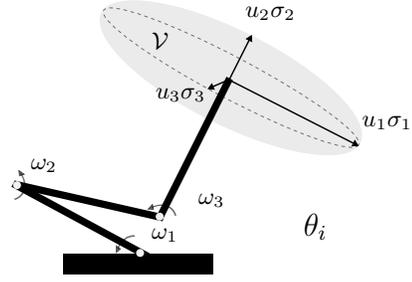
    \vspace{2mm}
\caption{Illustration of the manipulability ellipsoid of volume $\mathcal{V}$ for a manipulator end-effector at configuration $\boldsymbol{\theta}_i$.
    Larger axis lengths indicates higher mobility.}
    \label{fig:mnp_ellipsoid}
\end{figure}

Manipulability is a computationally tractable measure of the capacity for change in the pose of a robot given a specific joint configuration \cite{yoshikawa1985manipulability}.
It is associated with the ellipsoid defined by Eq.\ \eqref{eq:ellipsoid}, which is known as the \textit{manipulability ellipsoid} \cite{sciavicco2012modelling}.
The principal axes $\sigma_1\mathbf{u}_1, \sigma_2 \mathbf{u}_2 ... \sigma_m\mathbf{u}_m$ of this ellipsoid can be determined through singular value decomposition of $\mathbf{J} = \mathbf{U}\boldsymbol{\Sigma}\mathbf{V}^T$.
The manipulability measure of a given kinematic chain at $\boldsymbol{\theta}_i$ is defined as 
\begin{align}\label{eq:mnp}
\lambda = \sqrt{\det{\left(\mathbf{\mathbf{J}}\mathbf{J}^T\right)}} = \sigma_1 \sigma_2 \dots \sigma_k \dots \sigma_m,
\end{align}
and is proportional to the volume $\mathcal{V}$ of the manipulability ellipsoid \cite{sciavicco2012modelling}. 
The value $\sigma_k \geq 0 $ is the $k$-th largest singular value of $\mathbf{J}$, while $\mathbf{u}_{k}$ is the $k$-th column vector of $\mathbf{U}$.
A low manipulability corresponds to a low volume of the manipulability ellipsoid, inhibiting motion in the task space. An example of the manipulability ellipsoid of the end effector frame of a simple manipulator is depicted in Fig.\ \ref{fig:mnp_ellipsoid}. 

The gradient of Eq.\ \eqref{eq:mnp} can be calculated numerically, but it is also possible to derive its change with respect to the $j$-th joint of the configuration $\boldsymbol{\theta}_i$ using Jacobi's identity:
\begin{align}\label{eq:mnp_jacobian}
\frac{\partial \lambda}{\partial \theta_{i,j}} =  \frac{\lambda}{2}
											Tr\left( \left(\mathbf{J}\mathbf{J}^T\right)^{-1} 
											\left(
											\frac{\partial \mathbf{J}}{\partial \theta_{i,j}}\mathbf{J}^T 
											+ \mathbf{J}{\frac{\partial \mathbf{J} }{ \partial \theta_{i,j} }}^T \right)
											\right) \ .
\end{align}
The components of Eq.\ \eqref{eq:mnp_jacobian}, $\mathbf{J}$ and $\frac{\partial \mathbf{J}}{\partial\theta_{i,j}}$, can be calculated using geometrical methods \cite{hourtash2005kinematic}.

\subsection{Singularity Avoidance Likelihood}\label{subsection:likelihood}
Consider a kinematic chain and corresponding manipulability ellipsoid with a volume $\mathcal{V} \propto \lambda$, as shown in Fig.\ \ref{fig:mnp_ellipsoid}.
We define a minimum acceptable ellipsoid volume $\mathcal{V}_S \in \mathbb{R}_{+}$, and regard configurations resulting in a manipulability $\lambda < \lambda\left(\mathcal{V}_S\right)$ to be \textit{nearly singular} (labelled $S$).

Conversely, a high manipulability value does not guarantee that a configuration is not nearly singular, as an ellipsoid with one `degenerate' (i.e., of very small magnitude) axis may still have a large volume. The volume of manipulability ellipsoids for the chain is bounded by the value $\mathcal{V}_{max} < \infty$.
Assuming the axes $\sigma_1\textbf{u}_1, \sigma_2\textbf{u}_2 ... \sigma_m\textbf{u}_m$ are of an acceptable length for all such ellipsoids, we infer that configurations whose ellipsoid volume is sufficiently close to $\mathcal{V}_{max}$ are \emph{not} nearly singular (labelled $\bar{S}$).

Probabilistic inference provides an intuitive and efficient way to reason about the mapping between configurations and singularities. We define the likelihood of a given configuration not being nearly singular as:
\begin{align}\label{eq:singularity_likelihood}
L \left(\boldsymbol{\theta}_i | \bar{S}\right)  \propto p\left(\bar{S}| \boldsymbol{\theta}_i \right).
\end{align}
By modelling the distribution in Eq.\ \eqref{eq:singularity_likelihood}, we can optimize a chosen  trajectory prior to avoid singularities by maximizing the corresponding likelihood.
Our approach requires the distribution of Eq.\ \eqref{eq:singularity_likelihood} to take the form
\begin{align}\label{eq:singularity_likelihood2}
p\left(\bar{S}| \boldsymbol{\theta}_i \right)  =  \exp \{ -\frac{1}{2} \norm{ h_{\bar{S},i} }^2_{{\Sigma_{\bar{S}}}} \}.
\end{align}
Since the manipulability measure is proportional to a $m$-dimensional volume, its value may vary by several orders of magnitude throughout a trajectory.
In order to compensate for these large changes, we choose the cost $h_{\bar{S},i}$ to be logarithmic\footnote{Viewing $\Sigma_{\bar{S}}$ as a covariance matrix implies that $\frac{1}{\lambda}$ has a log-normal probability density function.},
\begin{align}\label{eq:mnp_likelihood}
h_{\bar{S},i} = \ln\left(\frac{\lambda_{max}}{\lambda}\right) ,
\end{align}
where the value $\lambda_{max}$ is the manipulability value at $\mathcal{V}_{max}$.
This improves cost gradient scaling by canceling out the $\lambda$ occurring in Eq.\  \eqref{eq:mnp_jacobian}.

\subsection{Trajectory Optimization}
A continuous-time trajectory is considered as a sample from a vector-valued Gaussian process (GP), $\boldsymbol{\theta}(t) \sim \mathcal{GP}(\boldsymbol{\mu}(t), \textbf{K}(t, t^{\prime})) $, with mean $\boldsymbol{\mu}(t)$ and covariance $ \textbf{K}(t, t^{\prime})$, generated by a linear time-varying stochastic differential equation (LTV-SDE)
\begin{equation}
\dot{\boldsymbol{\theta}}(t) = \textbf{A}(t) \boldsymbol{\theta}(t) + \boldsymbol{u}(t) + \boldsymbol{F}(t) \boldsymbol{w}(t),
\label{ltvsde}
\end{equation}
where $\boldsymbol{A}$ and $\boldsymbol{F}$ are system matrices, $\textbf{u}$ is a known control input and $\textbf{w}$ is generated by a white noise process.
In \cite{gpmp-ijrr}, such a representation is used to achieve efficient trajectory optimization. 
Assuming an exponential distribution for the trajectory prior $p\left(\boldsymbol{\theta}\right)$, finding a trajectory maximizing the likelihood in Eq.\ \eqref{eq:singularity_likelihood2} is a MAP problem:
\begin{align}\label{eq:MAP}
\boldsymbol{\theta}^* = \underset{\boldsymbol{\theta}}{\arg\max} \, p(\boldsymbol{\theta})\,p(\bar{S}\,|\,\boldsymbol{\theta}).
\end{align} 
The maximization in Eq.\ \eqref{eq:MAP} can be performed using methods such as Gauss-Newton or Levenberg-Marquardt.
This MAP trajectory optimization procedure can also be represented as inference on a factor graph \cite{kschischang2001factor} and solved efficiently using the SAM class of algorithms \cite{dong2016motion}.
Gradient information from additional states can be collected by GP interpolation, due to the Markovian property of the LTV-SDE describing the exactly sparse Gaussian process. Other factors representing likelihoods for constraints such as obstacle avoidance may be added in the same manner.

\section{Results} \label{section:results}
A nearly singular trajectory similar to the one shown in Fig.\ \ref{fig:IRL} can be generated by significantly extending the Universal Robots UR-10 arm\footnote{The UR-10 manipulator is available in our laboratory at the University of Toronto.} at the elbow joint, for example. The result is a loss of the capacity to move along the arm's extended axis. 
To demonstrate our approach to singularity avoidance, in this section we define an example motion planning problem where the final state is a Cartesian goal position and the trajectory prior is nearly singular throughout. The first scenario we describe places no constraints on arm movement, while the second scenario incorporates an obstacle in the arm's workspace.

We solve both problems using the GPMP2 algorithm \cite{dong2016motion}, comparing cases with and without the singularity avoidance factors described in Section \ref{section:singularities}.
The singularity avoidance factor covariance value in Eq.\  (\ref{eq:singularity_likelihood2}) is initialized as $\Sigma_{\bar{S}} = 10^{-4}$ with a GP power spectral density value of $\textbf{Q}_c = 10^3\,\mathbf{I}$. 
The position goal is defined as a factor on the final state of the factor graph in both cases, while singularity and collision \cite{gpmp-ijrr} factors are included in the optimized and interpolated states. We expect our method to increase the manipulability measure over the entire trajectory, `escaping' from the nearly singular prior while maintaining smoothness.

\subsubsection{Unconstrained workspace}

\begin{figure*}
	\centering
	\begin{subfigure}[]{0.32\textwidth}
		\includegraphics[width=\textwidth]{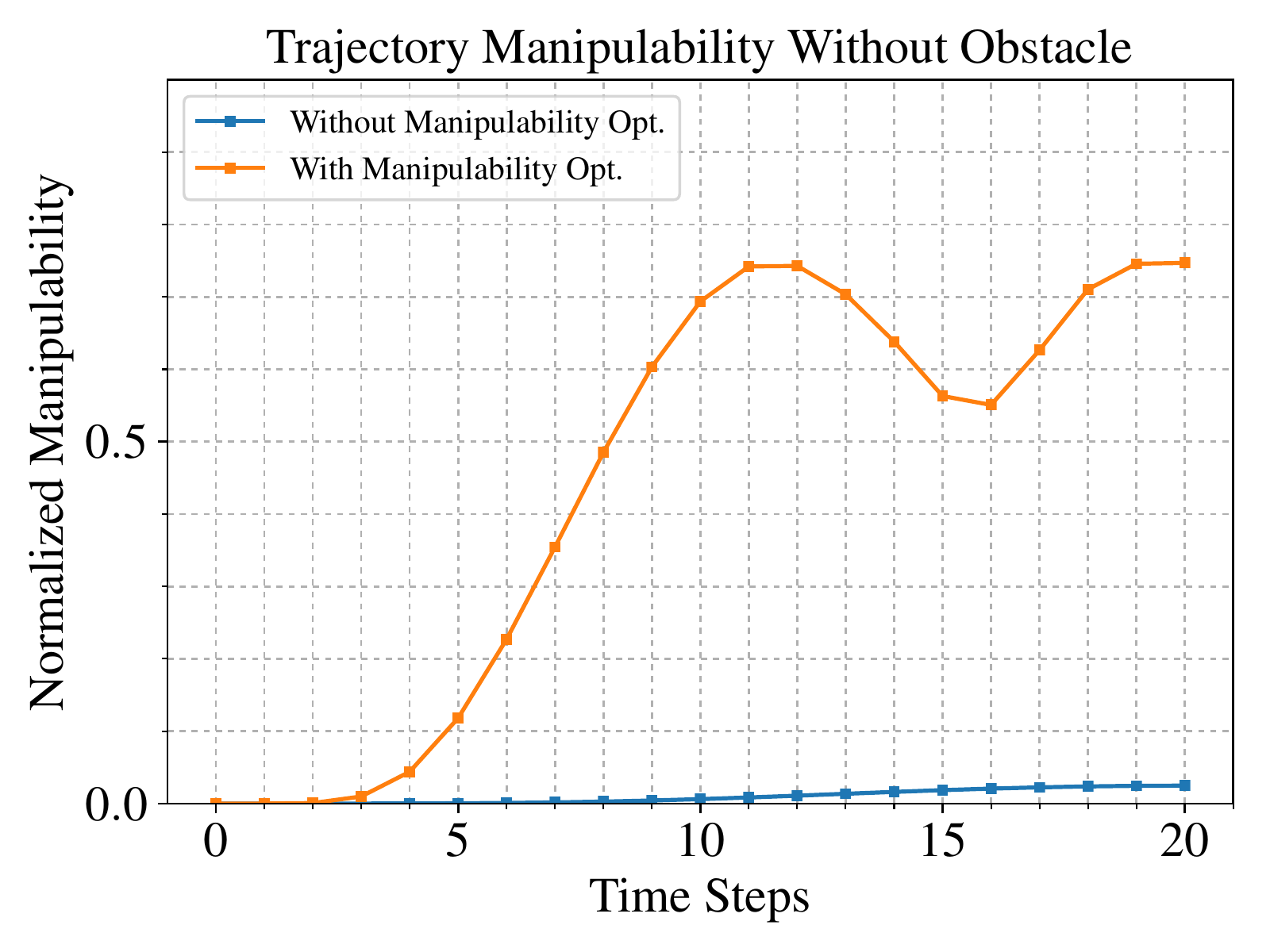}
		\caption{Normalized manipulability values.}
		\label{fig:norm_mnp_no_obs}
	\end{subfigure}
	\hfill
	\begin{subfigure}[]{0.32\textwidth}
		\resizebox {1\textwidth}{!}{
%
%

\begin{tikzpicture}

\begin{axis}[%
width=4.585in,
height=3.591in,
at={(0.769in,0.485in)},
scale only axis,
xmin=-0.613640076095255,
xmax=1.38635992390474,
tick align=outside,
xlabel style={font=\color{black}},
xlabel={\Large{x [m]}},
ymin=-0.596570996427332,
ymax=1.40342900357267,
ylabel style={font=\color{black}},
ylabel={\Large{y [m]}},
zmin=-1.27547979867006,
zmax=0.724520201329943,
zlabel style={font=\color{black}},
zlabel={\Large{z [m]}},
view={39.7999999999999}{14.4},
axis background/.style={fill=white},
title style={font=\bfseries},
title={},
axis x line*=bottom,
axis y line*=left,
axis z line*=left,
xmajorgrids,
ymajorgrids,
zmajorgrids,
legend style={legend cell align=left, align=left, draw=black}
]

\addplot3 [color=pyplotblue, line width=2.5pt, dashed]
 table[row sep=crcr] {%
0.577509041026266	0.215271030479692	-0.922226197134246\\
0.578554716593902	0.217555880137004	-0.921094710272546\\
0.581584387226822	0.224281622177191	-0.917747431735696\\
0.586379836699154	0.235269235081397	-0.912225770078129\\
0.59265613584344	0.250352537992485	-0.904538049645145\\
0.600072939579631	0.269369581409088	-0.894667092547946\\
0.608245365435909	0.292153431994713	-0.882577692563085\\
0.616754782403	0.31852298757134	-0.86822387431736\\
0.625159760271549	0.348274387098613	-0.851555840953845\\
0.633007329091434	0.381173516771802	-0.832526518350009\\
0.639844585754045	0.416950049637803	-0.811097606320294\\
0.645230568908903	0.4552933858515	-0.787245049176863\\
0.648748212993158	0.495850779260584	-0.760963841252969\\
0.650016095109067	0.538227841110441	-0.732272088815276\\
0.648699612152308	0.581991503526289	-0.701214259085219\\
0.644521176186131	0.626675406798462	-0.667863560311965\\
0.637268998195783	0.671787550377389	-0.632323414034199\\
0.626804046603632	0.716819924678071	-0.594728001490517\\
0.613064817396248	0.761259727221038	-0.555241889899135\\
0.596069634887141	0.804601670526343	-0.514058770050986\\
0.575916310881375	0.846360818186108	-0.471399363158303\\
0.552779117874178	0.886085345899517	-0.427508580859825\\
0.526903169690422	0.923368620028832	-0.382652046357608\\
0.498596440429994	0.957860018797781	-0.337112105565331\\
0.468219779382618	0.989273989020976	-0.291183473751829\\
0.436175386187636	1.017396929732	-0.24516867456416\\
0.402894289098909	1.04209161621925	-0.199373433899878\\
0.368823414322971	1.06329901479754	-0.15410219058537\\
0.334412843433958	1.08103748003623	-0.109653879265746\\
0.300103829254433	1.09539946182937	-0.0663181287118414\\
0.266318081625169	1.10654597004444	-0.0243720015852431\\
0.233448748911172	1.11469914146275	0.0159226195081568\\
0.201853416430777	1.12013332140965	0.0543209190227118\\
0.171849327745812	1.12316510745633	0.0905962538735642\\
0.143710917513294	1.12414280405078	0.124540449241702\\
0.11766963324279	1.12343570652586	0.155963283114519\\
0.0939159242501432	1.12142357430495	0.184691183029882\\
0.0726031937753124	1.11848657144534	0.210565181866439\\
0.0538534467290375	1.11499585398747	0.233438200874575\\
0.037764320463454	1.11130487428042	0.253171747564183\\
0.0244171565103008	1.10774135871897	0.269632134024117\\
0.0138857523488225	1.10459980286497	0.282686338585069\\
0.00624541697780678	1.10213422189309	0.292197651651481\\
0.00158193384870961	1.10055079951325	0.298021266537996\\
-5.77878439544932e-11	1.09999999995832	0.30000000003354\\
};
 \addlegendentry{{\small Without Manipulability Opt.}}
 
\addplot3 [color=pyplotred, line width=2.5pt, dashed]
 table[row sep=crcr] {%
0.577509040858943	0.21527103009443	-0.92222619729478\\
0.582858989072932	0.226871510644448	-0.914284587789618\\
0.59648408372089	0.259443514164931	-0.891757108437988\\
0.613615738107152	0.309624253506507	-0.856269411855912\\
0.629026262850284	0.373734536634605	-0.809450535976689\\
0.638143590774404	0.447601134241276	-0.753285965404374\\
0.637866754775851	0.526666030874951	-0.690222217912883\\
0.627006368079185	0.606319556502938	-0.623084451574867\\
0.606336431253428	0.682314485615973	-0.554925806621718\\
0.578330350402622	0.751122655397923	-0.488917051974575\\
0.546705802777492	0.810144688850037	-0.428325239255591\\
0.515900618449186	0.857725606654159	-0.376566524119291\\
0.486856853736223	0.895664402291682	-0.333060087063794\\
0.457839658447637	0.92721079380518	-0.294427851627446\\
0.429258913684552	0.953354744212961	-0.259993644431873\\
0.401269968142781	0.974910109321538	-0.229021763299309\\
0.373842714543591	0.992528621968228	-0.200734592715298\\
0.346814150544867	1.00671610831187	-0.174325831402561\\
0.319928622723211	1.01784730340473	-0.148971797621839\\
0.292869330868313	1.0261769089435	-0.123842482652456\\
0.265284161997084	1.03184544999998	-0.098113522247926\\
0.236808591418945	1.03487922304348	-0.0709799473941035\\
0.207088247466446	1.03518432288335	-0.0416723970235795\\
0.178690327031331	1.03141580083961	-0.0142149153004561\\
0.154248771604789	1.02297830069339	0.00766978684429982\\
0.133444676148443	1.01065513234759	0.0252037274837672\\
0.115879408150942	0.995082268557809	0.0396294585966211\\
0.10110680168332	0.976805144686706	0.0521749052020781\\
0.088655103491431	0.956319968137318	0.0640221327491601\\
0.0780429960157276	0.934097655176483	0.0762810267906658\\
0.0687925129775816	0.910590162680362	0.0899697529934833\\
0.0604406036539653	0.886220207634327	0.10600404533534\\
0.0525505977760631	0.86135609868554	0.125196831139301\\
0.0447249248817106	0.836273552369332	0.148268490715272\\
0.0372261752558702	0.822190143015109	0.170771524319002\\
0.0298857063503066	0.828372771027774	0.189755427143798\\
0.0224357547459708	0.850976959162194	0.207249687169547\\
0.0152510501423896	0.885331685072172	0.224227745386066\\
0.00895595745829435	0.926528808685979	0.240836125256507\\
0.00411110006985109	0.969870366273187	0.256664984433334\\
0.0010051004993249	1.01121442757414	0.271013121256985\\
-0.000438121016185081	1.04719447907878	0.283112797575745\\
-0.000640991108664273	1.07525605939049	0.292287920933394\\
-0.000260004998585812	1.09345553965323	0.298031067932712\\
-6.47155763283669e-08	1.09999089714434	0.300001087707757\\
};
 \addlegendentry{{\small With Manipulability Opt.}}

\addplot3 [color=pyplotgreen, line width=3.0pt]
 table[row sep=crcr] {%
0	0	0.1273\\
0.165606006506759	0.16560600657973	-0.43811427395636\\
0.320843065896148	0.320843066037087	-0.96663125643012\\
0.43676685867879	0.204919273190722	-0.966631256463745\\
0.512288758705089	0.280441173196162	-0.922141086358584\\
0.577509040858943	0.21527103009443	-0.92222619729478\\
};
 \addlegendentry{{\small Starting}}

\addplot3 [color=red, draw=none, mark size=3.5pt, mark=square*, mark options={solid, fill=red, red}]
 table[row sep=crcr] {%
-6.47155763283669e-08	1.09999089714434	0.300001087707757\\
};
 \addlegendentry{{\small Goal}}
 
\addplot3 [color=black, draw=none, mark size=3.5pt, mark=*, mark options={solid, black}]
 table[row sep=crcr] {%
0	0	0.1273\\
0.165606006506759	0.16560600657973	-0.43811427395636\\
0.320843065896148	0.320843066037087	-0.96663125643012\\
0.43676685867879	0.204919273190722	-0.966631256463745\\
0.512288758705089	0.280441173196162	-0.922141086358584\\
};

\addplot3 [color=pyplotred, line width=3.0pt]
 table[row sep=crcr] {%
0	0	0.1273\\
0.00893925812497275	0.601551525514065	0.239549061536729\\
0.0147335658869615	0.991469199381184	-0.179327917543782\\
0.178656467371579	0.989033249902145	-0.179327917577407\\
0.180080428614462	1.08485619789604	-0.114501804226862\\
0.207088247466446	1.03518432288335	-0.0416723970235795\\
};

\addplot3 [color=black, draw=none, mark size=3.5pt, mark=*, mark options={solid, black}]
 table[row sep=crcr] {%
0	0	0.1273\\
0.00893925812497275	0.601551525514065	0.239549061536729\\
0.0147335658869615	0.991469199381184	-0.179327917543782\\
0.178656467371579	0.989033249902145	-0.179327917577407\\
0.180080428614462	1.08485619789604	-0.114501804226862\\
};

\addplot3 [color=pyplotblue, line width=3.0pt]
 table[row sep=crcr] {%
0	0	0.1273\\
0.126078145416058	0.446431420029286	-0.271883276777827\\
0.269716426993858	0.955041709158722	-0.491453101337231\\
0.42748646036781	0.910485362246278	-0.491453101370856\\
0.438764836485957	0.950421081849132	-0.383451126704002\\
0.526903169690422	0.923368620028832	-0.382652046357608\\
};

\addplot3 [color=black, draw=none, mark size=3.5pt, mark=*, mark options={solid, black}]
 table[row sep=crcr] {%
0	0	0.1273\\
0.126078145416058	0.446431420029286	-0.271883276777827\\
0.269716426993858	0.955041709158722	-0.491453101337231\\
0.42748646036781	0.910485362246278	-0.491453101370856\\
0.438764836485957	0.950421081849132	-0.383451126704002\\
};

\end{axis}
\end{tikzpicture}%
}		
		\caption{Simulated trajectories.}
		\label{fig:mnp_no_obs_3d}
	\end{subfigure}
	\hfill
	\begin{subfigure}[]{0.32\textwidth}
		\includegraphics[width=\textwidth]{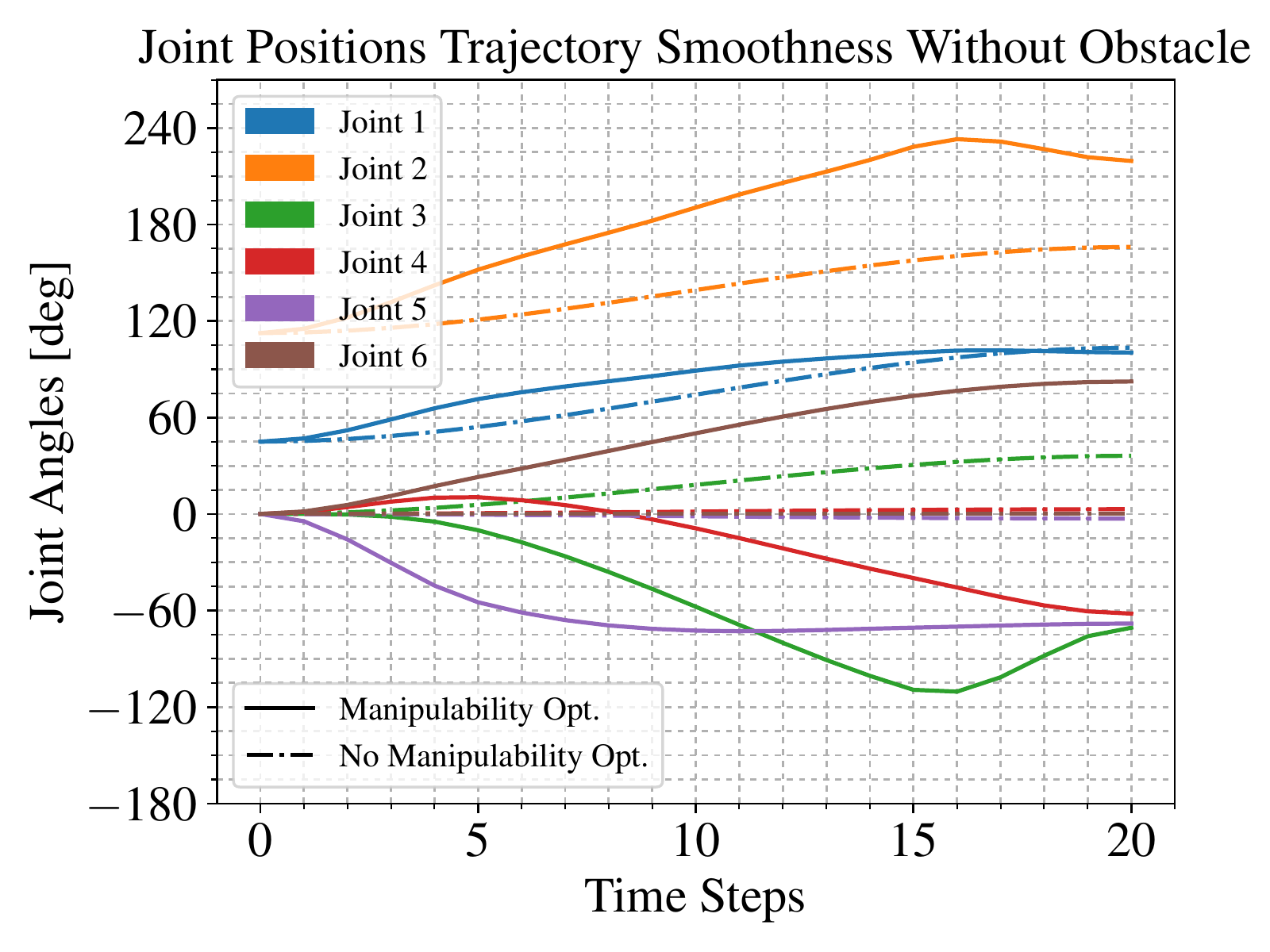}
		\caption{Joint trajectories.}
		\label{fig:mnp_no_obs_joint}
	\end{subfigure}
	\caption{Experimental results for both the prior and optimized trajectories, without collision avoidance. The values are normalized to the maximum observed manipulability value of 2.93.}
\end{figure*}
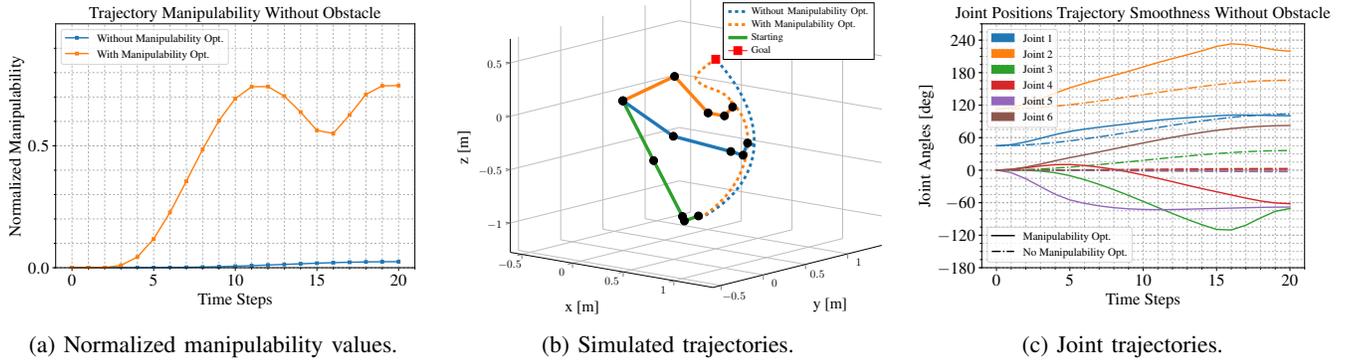

Fig.\ \ref{fig:norm_mnp_no_obs} shows that our method results in a significant increase in the manipulability measure, as defined by Eq.\ (\ref{eq:mnp}), over the entire trajectory.
The standard GPMP2 method produces a smooth solution, maintaining a minimum distance from the initial trajectory as governed by the GP prior factors.
The change in trajectory achieved by our method is seen in Fig.\ \ref{fig:mnp_no_obs_3d}---the arm is (visibly) less extended and regains its movement capability along the previously degenerate axis.
The GP prior factors also maintain smoothness in our solution (with manipulability optimization), as seen in Fig.\ \ref{fig:mnp_no_obs_joint}, where there are no sudden jumps or changes in the joint values.

\subsubsection{Obstacle in workspace}

\begin{figure*}
	\centering
	\begin{subfigure}[]{0.32\textwidth}
		\includegraphics[width=\textwidth]{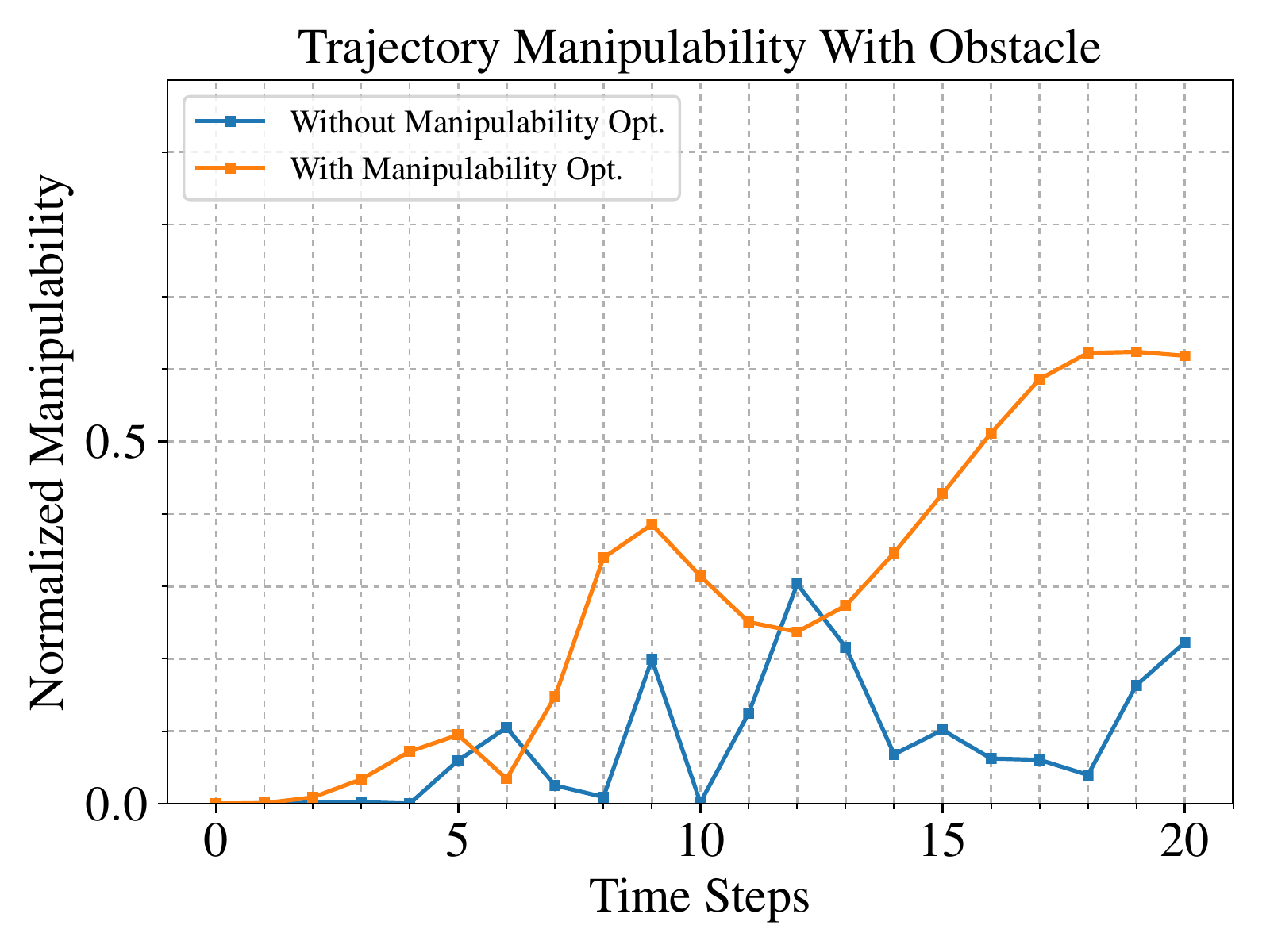}
		\caption{Normalized manipulability values.}
		\label{fig:norm_mnp_with_obs}
	\end{subfigure}
	\hfill
	\begin{subfigure}[]{0.28\textwidth}
		\resizebox {1\textwidth}{!}{\input{figures/fig2.tex}}			
		\caption{Simulated trajectories.}
		\label{fig:mnp_with_obs_3d}
	\end{subfigure}
	\hfill
	\begin{subfigure}[]{0.32\textwidth}
		\includegraphics[width=\textwidth]{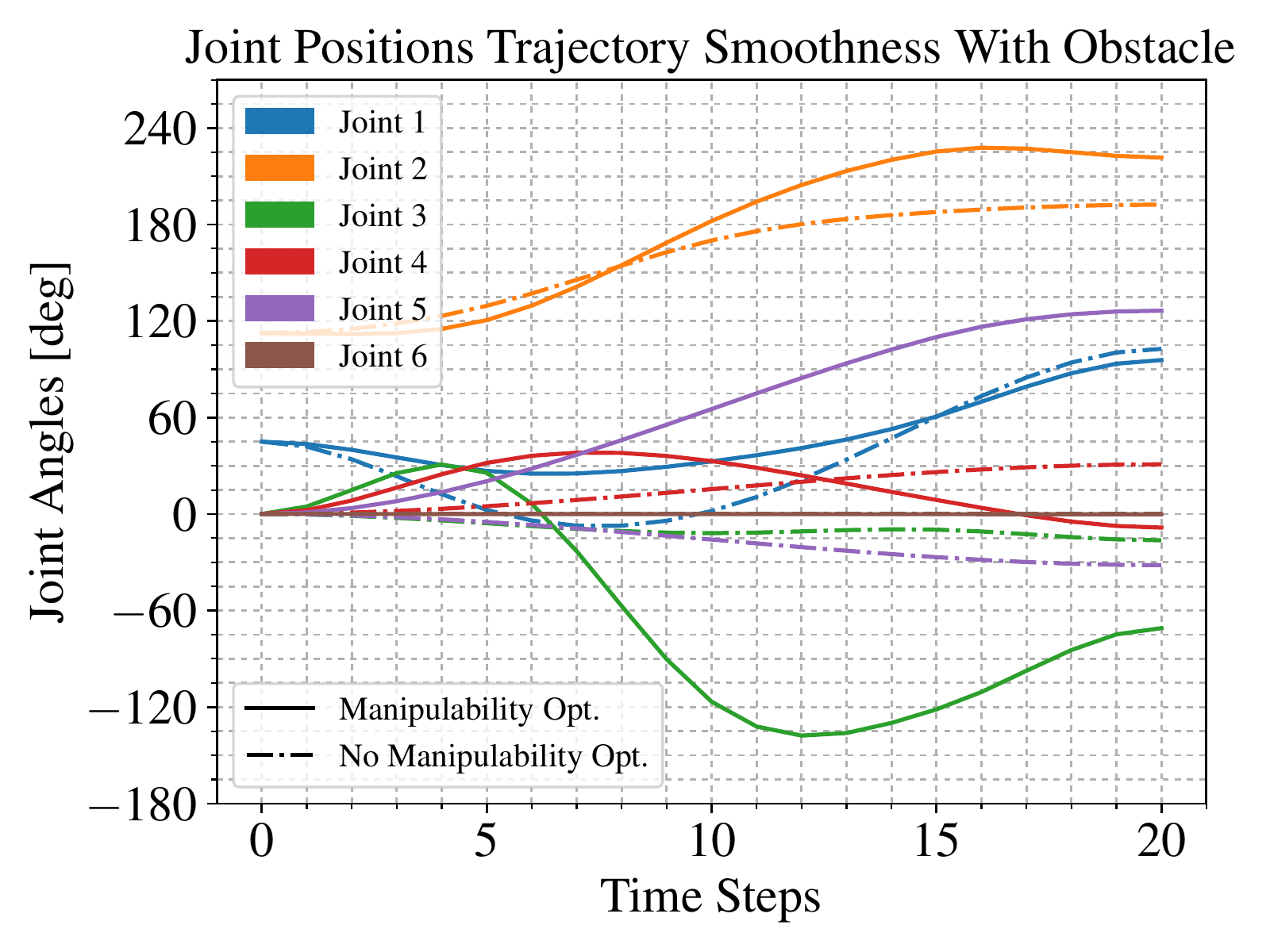}
		\caption{Joint trajectories.}
		\label{fig:mnp_with_obs_joint}
	\end{subfigure}
	\caption{Experimental results for both the prior and optimized trajectory with collision avoidance included. The values are normalized to the maximum observed manipulability value of 2.93.}
\end{figure*}

\begin{figure}
	\centering
	\begin{subfigure}[]{0.78\columnwidth}
		\includegraphics[width=\textwidth]{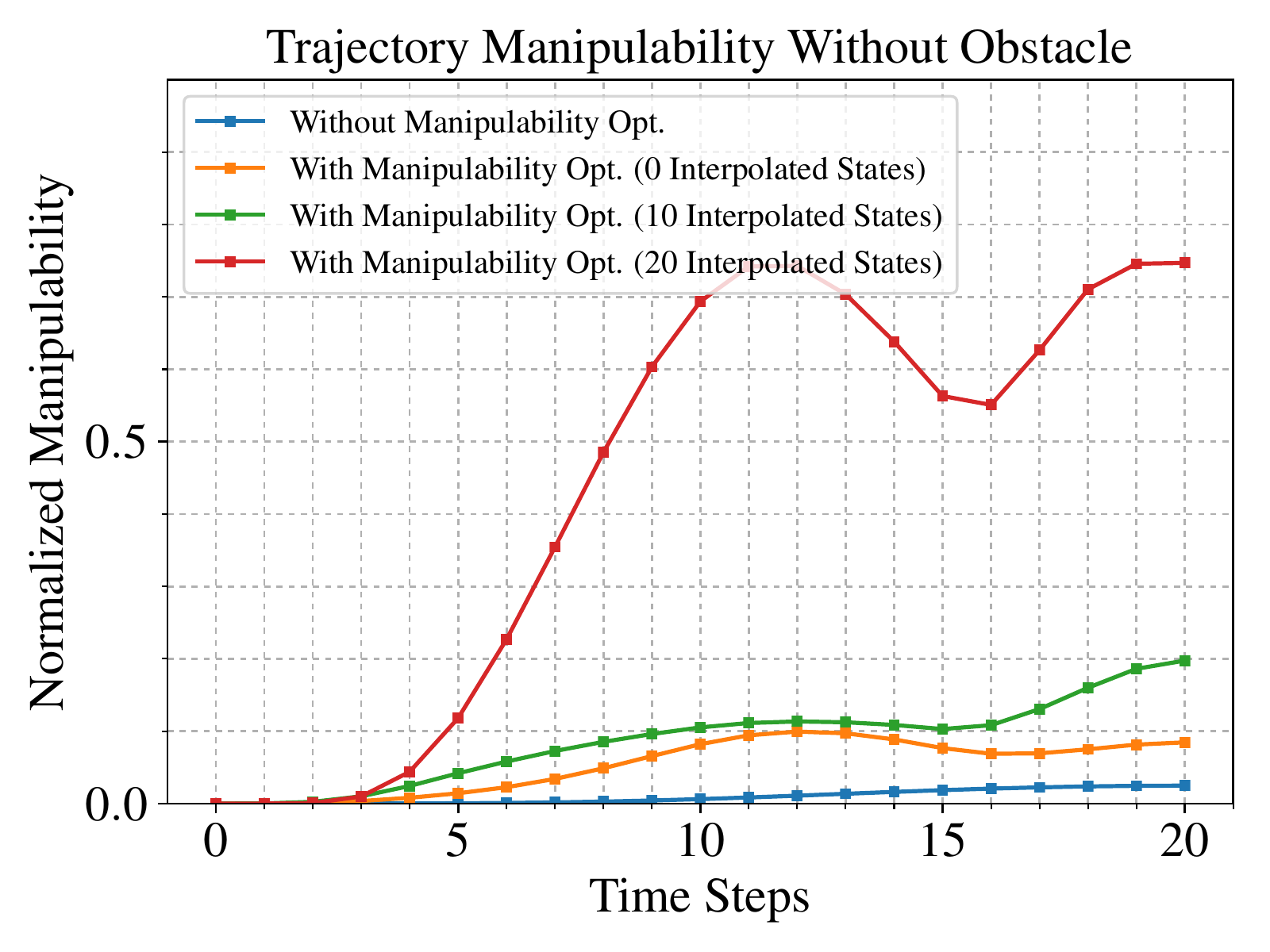}	
	\end{subfigure}
	\begin{subfigure}[]{0.79\columnwidth}
		\includegraphics[width=\textwidth]{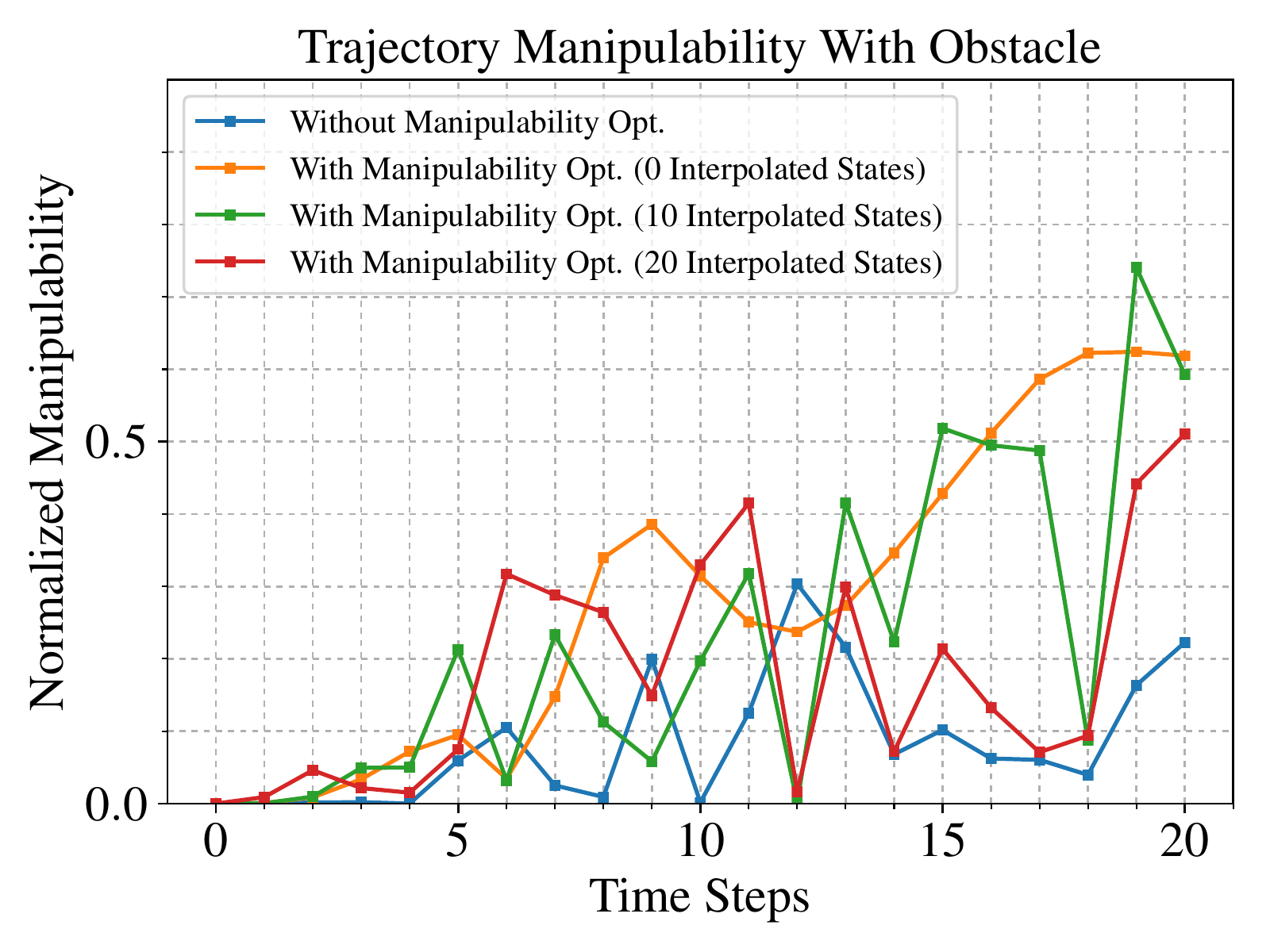}
	\end{subfigure}
	\caption{Increase of manipulability with a larger number of interpolated states. Values are normalized to the highest observed value.}
	\label{fig:inteprolation_improvement}
\end{figure}

In order to ensure that multiple factors can be successfully optimized simultaneously, we repeated the test above while placing a table-shaped obstacle in the workspace (hence necessitating the use of collision avoidance factors). We choose singularity and collision avoidance factor covariance values of $\Sigma_{\bar{S}} = 10^{-2}$ and $\boldsymbol{\Sigma}_{obs} = 10^{-3}\mathbf{I}$, with a GP power spectral density value of $\textbf{Q}_c = 10^3 \mathbf{I}$.

In this case, the trajectory prior passed through the table, making the initial collision cost non-zero.
Fig.\ \ref{fig:mnp_with_obs_3d} shows both resulting trajectories in the collision avoidance scenario as feasible solutions, with our method achieving a significant improvement in manipulability, as can be seen in Fig.\ \ref{fig:norm_mnp_with_obs}.
As demonstrated by Fig.\ \ref{fig:mnp_with_obs_joint}, both trajectories maintain the smoothness imposed by the prior. Note that the final trajectory produced by the standard GPMP2 method  remains nearly singular, despite avoiding the obstacle. 
Our method finds a solution with a visibly larger capacity for further motion in the task space, by bending the arm at the elbow to pass directly in front of the table. This is instead of the more conservative trajectory that avoids the table but compromises the arm’s manipulability, maintaining a near-singular (overextended) configuration throughout the motion.

An interesting result is shown in Fig.\ \ref{fig:inteprolation_improvement}, which characterizes how increasing the number of interpolation states affects the manipulability over the trajectory.
In the first scenario, where no obstacle is present, increasing the number of interpolation states improves the results by a large margin, whereas in the second scenario, with an obstacle, the need to avoid collisions offsets this manipulability improvement.
A possible explanation is that propagating interpolated gradients for both collision and singularity costs to support states fails to provide clear gradient information for optimizing the support states.
Nonetheless, we observe that in all cases, even when avoiding collision, the manipulability values are higher than those in the prior.

\section{Conclusion and Future Work} 
\label{section:future_work_conclusion}
In this paper we presented a novel method for avoiding singular configuration during planning and replanning using a continuous-time Gaussian process trajectory representation.
This allowed us to formulate the problem in a maximum likelihood framework, which, for our representation, is solvable using a MAP estimator. Initial results show significant manipulability improvements, which we have shown in our analyses.

Preliminary tests demonstrate that our method can be used to achieve rapid replanning with the iSAM2 algorithm, however we leave a full experimental treatment as future work. Exploring the properties manipulability ellipsoids when taking into account configuration uncertainties could prove to be an interesting future research direction. This may provide an alternative formulation of the likelihood function.

\balance
\bibliographystyle{IEEEcaps}
\bibliography{refs}

\begin{thebibliography}{10}
\providecommand{\url}[1]{#1}
\csname url@rmstyle\endcsname
\providecommand{\newblock}{\relax}
\providecommand{\bibinfo}[2]{#2}
\providecommand\BIBentrySTDinterwordspacing{\spaceskip=0pt\relax}
\providecommand\BIBentryALTinterwordstretchfactor{4}
\providecommand\BIBentryALTinterwordspacing{\spaceskip=\fontdimen2\font plus
\BIBentryALTinterwordstretchfactor\fontdimen3\font minus
  \fontdimen4\font\relax}
\providecommand\BIBforeignlanguage[2]{{%
\expandafter\ifx\csname l@#1\endcsname\relax
\typeout{** WARNING: IEEEtran.bst: No hyphenation pattern has been}%
\typeout{** loaded for the language `#1'. Using the pattern for}%
\typeout{** the default language instead.}%
\else
\language=\csname l@#1\endcsname
\fi
#2}}

\bibitem{sciavicco2012modelling}
L.~Sciavicco and B.~Siciliano, \emph{Modelling and Control of Robot
  Manipulators}, ser. Advanced Textbooks in Control and Signal
  Processing.\hskip 1em plus 0.5em minus 0.4em\relax Berlin, Heidelberg:
  Springer Science \& Business Media, 2012.

\bibitem{yoshikawa1985manipulability}
T.~Yoshikawa, ``Manipulability of Robotic Mechanisms,'' \emph{The International
  Journal of Robotics Research}, vol.~4, no.~2, pp. 3--9, 1985.

\bibitem{chiaverini1997singularity}
S.~Chiaverini, ``Singularity-Robust Task-Priority Redundancy Resolution for
  Real-Time Kinematic Control of Robot Manipulators,'' \emph{IEEE Transactions
  on Robotics and Automation}, vol.~13, no.~3, pp. 398--410, 1997.

\bibitem{guilamo2006manipulability}
L.~Guilamo, J.~Kuffner, K.~Nishiwaki, and S.~Kagami, ``Manipulability
  Optimization for Trajectory Generation,'' in \emph{Proceedings of the IEEE
  International Conference on Robotics and Automation (ICRA'06)}, May 2006, pp.
  2017--2022.

\bibitem{dufour2017integrating}
K.~Dufour and W.~Suleiman, ``On Integrating Manipulability Index into Inverse
  Kinematics Solver,'' in \emph{Proceedings of the IEEE/RSJ International
  Conference on Intelligent Robots and Systems (IROS'17)}, September 2017, pp.
  6967--6972.

\bibitem{barfoot2014batch}
T.~D. Barfoot, C.~H. Tong, and S.~S{\"a}rkk{\"a}, ``Batch Continuous-Time
  Trajectory Estimation as Exactly Sparse Gaussian Process Regression,'' in
  \emph{Proceedings of Robotics: Science and Systems (RSS'14)}, 2014.

\bibitem{gpmp-ijrr}
M.~Mukadam, J.~Dong, X.~Yan, F.~Dellaert, and B.~Boots, ``Continuous-time
  Gaussian Process Motion Planning via Probabilistic Inference,'' \emph{arXiv
  preprint arXiv:1707.07383}, 2017.

\bibitem{dong2016motion}
J.~Dong, M.~Mukadam, F.~Dellaert, and B.~Boots, ``Motion Planning as
  Probabilistic Inference using Gaussian Processes and Factor Graphs.'' in
  \emph{Proceedings of Robotics: Science and Systems (RSS'16)}, 2016.

\bibitem{dellaert2006square}
F.~Dellaert and M.~Kaess, ``Square Root SAM: Simultaneous Localization and
  Mapping via Square Root Information Smoothing,'' \emph{The International
  Journal of Robotics Research}, vol.~25, no.~12, pp. 1181--1203, 2006.

\bibitem{hourtash2005kinematic}
A.~Hourtash, ``The Kinematic Hessian and Higher Derivatives,'' in
  \emph{Proceedings of the IEEE International Symposium on Computational
  Intelligence in Robotics and Automation (CIRA'05)}, June 2005, pp. 169--174.

\bibitem{kschischang2001factor}
F.~R. Kschischang, B.~J. Frey, and H.-A. Loeliger, ``Factor Graphs and the
  Sum-Product Algorithm,'' \emph{IEEE Transactions on information theory},
  vol.~47, no.~2, pp. 498--519, 2001.

\end{thebibliography}

\end{document}